\documentclass[final]{cvpr}

\usepackage{times}
\usepackage{epsfig}
\usepackage{graphicx}
\usepackage{amsmath}
\usepackage{amssymb}
\usepackage{multirow}
\usepackage{graphicx}
\usepackage{booktabs}
\usepackage{subfigure}
\usepackage{xcolor}
\usepackage{colortbl,booktabs}
\usepackage{amsfonts}
\usepackage[keeplastbox]{flushend}
\usepackage{bm}

\usepackage[pagebackref=true,breaklinks=true,colorlinks,bookmarks=false]{hyperref}

\def\cvprPaperID{3342} 
\def\confYear{CVPR 2021}

\begin{document}

\title{Self-Promoted Prototype Refinement for Few-Shot Class-Incremental Learning}


\author{Kai~Zhu\textsuperscript{\rm 1}\footnotemark[1]\qquad~Yang~Cao\textsuperscript{\rm 1}\footnotemark[1]\qquad~Wei~Zhai\textsuperscript{\rm 1}\qquad~Jie~Cheng\textsuperscript{\rm 2}\qquad~Zheng-Jun~Zha\textsuperscript{\rm 1}\footnotemark[2]\\
		{\textsuperscript{\rm 1} University of Science and Technology of China}  \qquad
		{\textsuperscript{\rm 2} Huawei Technologies Co. Ltd.} \\
		\small{\texttt{\{zkzy@mail., forrest@, wzhai056@mail.\}ustc.edu.cn}} \qquad
		\small{\texttt{jiecheng2009@google.com}} \qquad
		\small{\texttt{zhazj@ustc.edu.cn}}
	}



\maketitle
\pagestyle{empty}  
\thispagestyle{empty}
\renewcommand{\thefootnote}{\fnsymbol{footnote}}
\footnotetext[1]{Co-first Author}
\footnotetext[2]{Corresponding Author}

\begin{abstract}
   Few-shot class-incremental learning is to recognize the new classes given few samples and not forget 
   the old classes. It is a challenging task since representation optimization and prototype reorganization 
   can only be achieved under little supervision. To address this problem, we propose a novel incremental prototype 
   learning scheme. Our scheme consists of a random episode selection strategy that adapts the feature 
   representation to various generated incremental episodes to enhance the corresponding extensibility, and a 
   self-promoted prototype refinement mechanism which strengthens the expression ability of the new classes by explicitly 
   considering the dependencies among different classes. Particularly, a dynamic relation projection 
   module is proposed to calculate the relation matrix in a shared embedding space and leverage it as the factor for bootstrapping 
   the update of prototypes. Extensive experiments on three benchmark datasets demonstrate the 
   above-par incremental performance, outperforming state-of-the-art methods by a margin of $13\% $, $17\% $ 
   and $11\% $, respectively.
\end{abstract}

\begin{figure}[t]
   \begin{center}
      \includegraphics[width=0.9\linewidth]{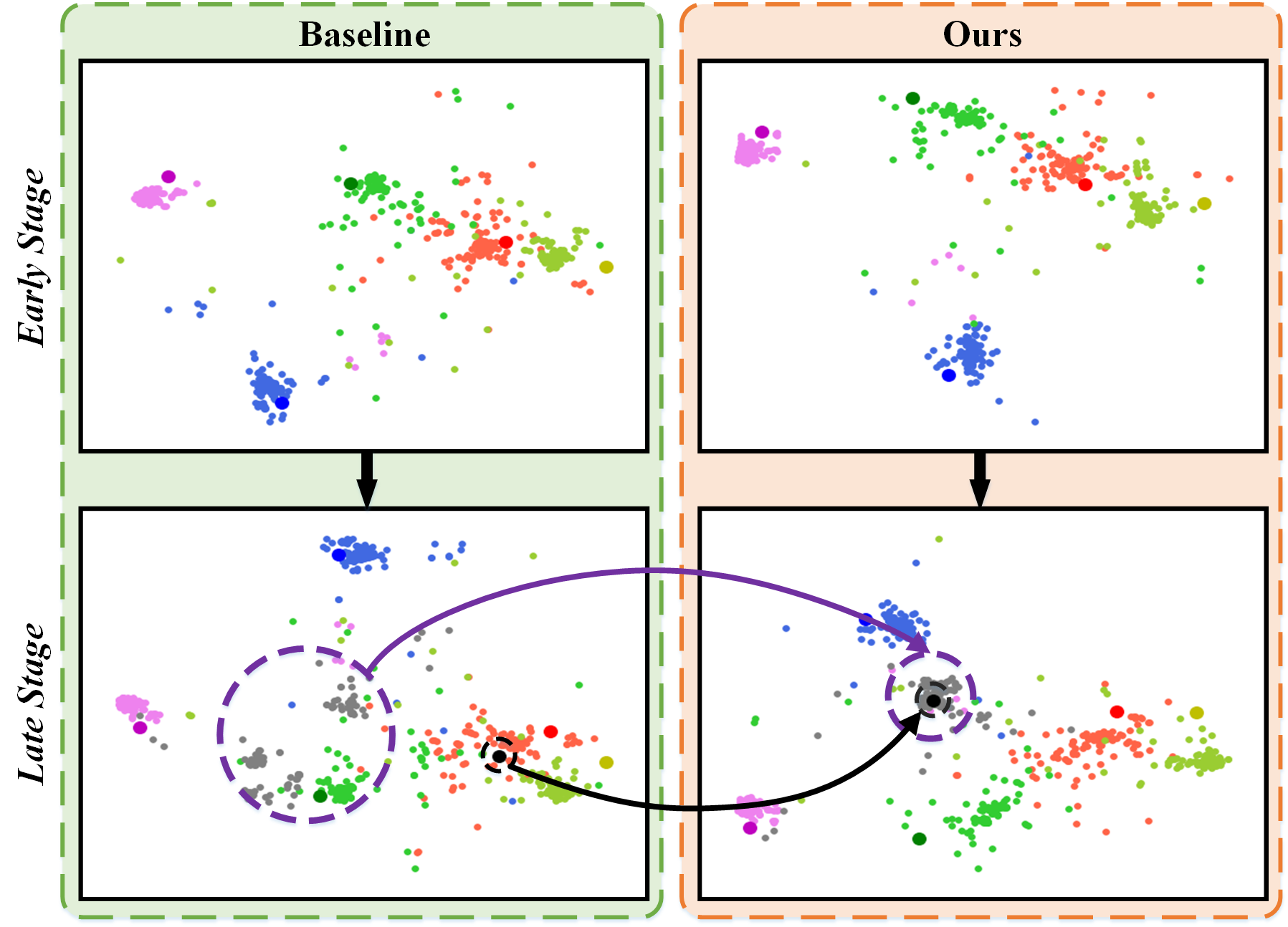}
   \end{center}
      \caption{The t-SNE \cite{maaten2008visualizing} results in different methods at two stages. The initial deep representations obtained by 
      typical incremental learning method \cite{rebuffi2017icarl, Belouadah2019IL2MCI} and our proposed method are visualized in the upper row (60 classes 
      used, and 5 classes visualized in color). In each colored class, deep-color points are learnable prototypes, 
      and light-color ones show the distribution of real data. The lower row shows the refined representations
      and prototypes of each class after the increment of the gray class. Compared 
      with previous method, (1) the representations of the incremental classes are more clustered (regions 
      circled in violet dotted lines), (2) and their corresponding prototypes are more discriminative, 
      where the incremental prototype visualized in black is representative and no longer 
      confused with that in red color.}
   \label{fig:first}
\end{figure}

\section{Introduction}
Currently, deep convolutional neural networks \cite{Simonyan2015VeryDC, he2016deep, zhang2020empowering} have made significant breakthroughs in a large number of recognition tasks.
When the class is given in advance and the sample is sufficient, we can get a good recognition model by typical supervised 
learning. In practice, however, we are likely to encounter new classes that were not seen before in continual data stream, and 
need to add them into the recognition tasks, which forms the problem of class-incremental learning (CIL) \cite{rebuffi2017icarl}. 

In this case, it is both 
time consuming and computationally expensive to retrain the model on all the old and new data. And in many cases the old 
data may not be available, due to data privacy or limited storage. A common solution is to fine-tune old models with new data, 
but it may arise the problem of catastrophic forgetting. To this end, 
recent learning-based approaches present to maintain the representation space for the old 
classes by preserving memories of old classes (e.g., examplar \cite{rebuffi2017icarl}) and introducing 
various distillation losses, and then reconstruct classifiers (e.g., fully connected layer \cite{wu2019large}, learnable 
prototypes \cite{hou2019learning}) in different ways to correct their preference for new classes.

However, existing methods assume that new class samples are available in large quantities, while the 
incremental classes are usually atypical and the sample size is small in practical applications. For example, 
in industrial visual inspection tasks, with the continuous progress of production, new classes of defects often 
appear due to equipment wear and other reasons. These defect samples may not only be essentially different from 
the old samples, but also small in number. It brings great difficulties to the recognition task, as 
representation optimization and prototype reorganization brought by new classes are hard to complete under little supervision.  
This paper focuses on this ability of incrementally 
learning new classes from few samples, which is called few-shot class-incremental learning (FSCIL \cite{tao2020few}). 

A natural idea for FSCIL task is to directly apply existing incremental learning methods to solve the problem, but experimental 
results show that this way results in a dramatic drop in performance. Our analysis suggests that this is mainly 
due to the following two reasons. 
First, the initial deep representation space used for CIL is relatively compact, which is 
conducive to classification of existing classes, but it lacks extensibility for FSCIL. 
In FSCIL, due to the insufficiency of incremental class samples, there is no enough supervision at 
each stage to participate in the classification and distillation process. Therefore, it cannot promote the expansion 
of the representation space as the existing incremental learning methods do.
In addition, the small number of new class samples is not sufficient to learn discriminative classifier for new classes 
while maintaining the performance on old classes. As shown in Fig. \ref{fig:first}, the representation 
space extended by typical incremental learning approach \cite{rebuffi2017icarl, castro2018end} is underrepresented, such that the 
new classes usually exhibit insufficient aggregation compared to the old ones. Also, due to the insufficiency of 
new class samples, the prototypes used for classification are prone to be confused with other classes after 
incremental learning, which greatly deteriorates subsequent tasks.

To address this problem, we propose an incremental prototype learning scheme to explicitly learn an 
extensible feature representation, and thus facilitate subsequent incremental tasks. The scheme is mainly manifested in 
two aspects. First, we adopt the random episode selection strategy (RESS) to enhance the extensibility of 
feature representation 
by forcing features adaptive to various randomly simulated incremental processes. Secondly, we introduce 
a self-promoted prototype refinement mechanism (SPPR) to update the existing prototypes by utilizing the relation matrix 
between representations of the new class samples and the old class prototypes. This enhances the expressiveness 
of the new classes while retaining the relational characteristics among the old classes. Particularly, 
a novel module called dynamic relation projection is proposed to map the representation 
of the new class samples and the prototype of the old classes into the same embedding space, and calculate a projection 
matrix between them by using the distance metric of the two embeddings in the space. 
We take the matrix as the weight of prototype refinement to guide the dynamic change of the prototype toward 
maintaining the existing knowledge and enhancing the discriminability of the new class. To 
demonstrate the superiority of our method, we conducted comparative experiments with existing 
few-shot class-incremental and typical class-incremental methods on three datasets CIFAR-100, MiniImageNet and CUB200. 
We achieved the best results against the state-of-the-art methods, leading by $13\%$ , $17\% $, and $11\% $, respectively.

Our main contributions are as follows:

1.	An incremental prototype learning scheme is proposed for few-shot class-incremental learning, in which a 
randomly episodic training is accomplished by a self-promoted prototype refinement mechanism, resulting in an 
extensible feature representation.

2.	A novel dynamic relation projection module is proposed, which uses the relational metric between old class 
prototypes and new class samples to constrain the update of prototypes during training and test. 

3. Extensive experiments on benchmark CIFAR-100, MiniImageNet and CUB200 datasets demonstrate the superiority of 
our proposed method over the state-of-the-art.

\begin{figure*}
   \begin{center}
      \includegraphics[width=0.87\linewidth]{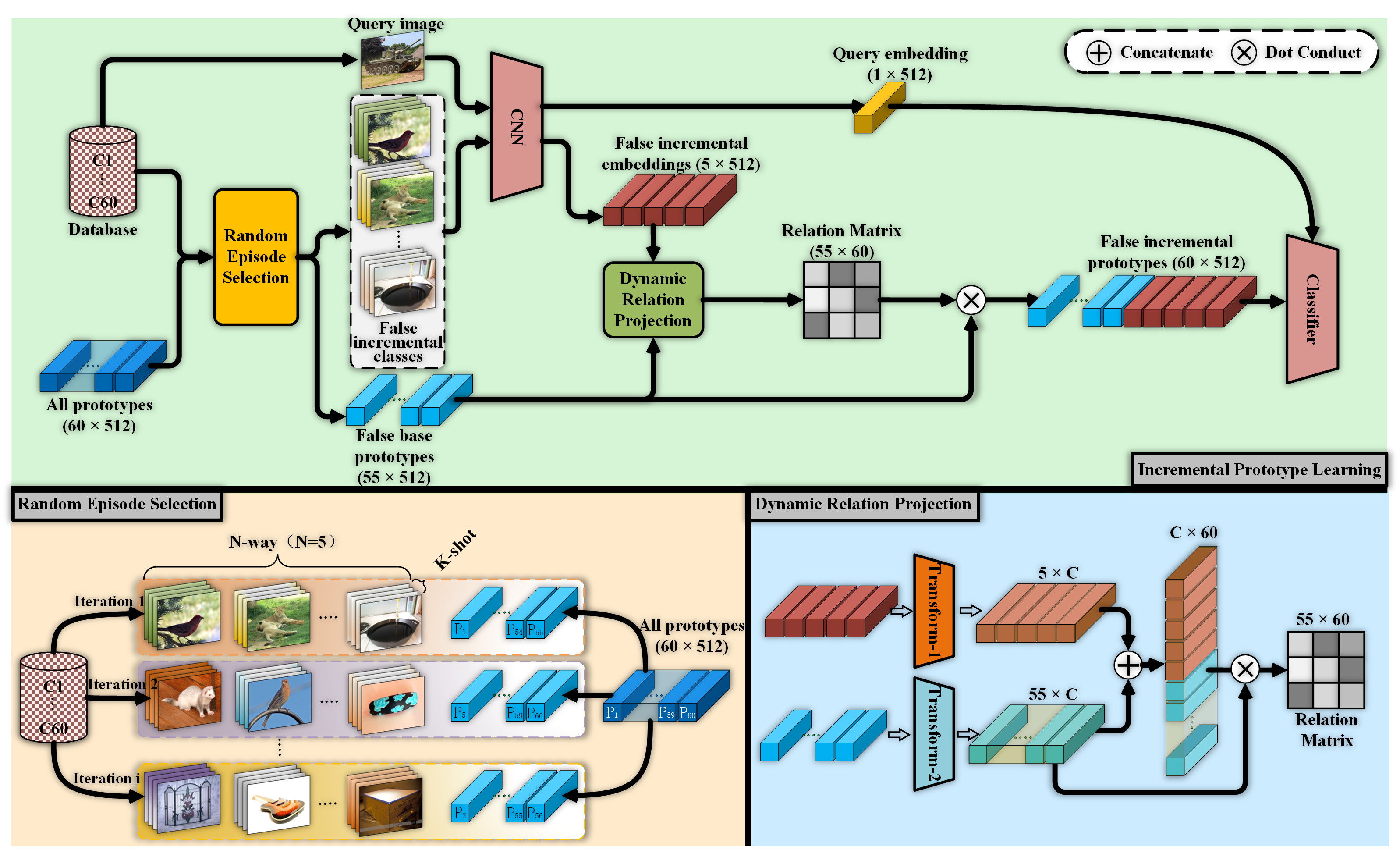}
   \end{center}
   \caption{Our incremental prototype learning scheme for few-shot class-incremental learning. (a) Overview 
   of our scheme. (b) Random episode selection strategy. (c) Dynamic relation projection module.}
   \label{fig:model}
\end{figure*}

\section{Related Work}
\subsection{Incremental Learning}
Although deep neural networks have shown excellent performance in many individual tasks, it still remains 
a substantial challenge to learn different tasks in sequence. Thus incremental learning \cite{Li2018LearningWF} continues to receive 
much attention. According to whether the task identity is informed or needs to be inferred, incremental 
learning can be divided into three categories \cite{Ven2019ThreeSF}: task-incremental learning, domain-incremental learning and 
class-incremental learning (CIL). CIL is the most challenging and the closest to 
the needs of practical applications. Therefore, a large number of related studies emerge in recent years.
iCARL \cite{rebuffi2017icarl} preserves valuable samples of the old classes  
called exemplars with a limited capacity, and designs a set of strategies for selecting and updating them. It 
decouples the representation learning and classification by an examplar-rehearsed knowledge distillation and a 
nearest-mean-of-exemplars rule. Most of the subsequent works follow this framework and makes corresponding improvements. 

NCM \cite{hou2019learning} incorporates cosine normalization, less-forget constraint and inter-class separation, to mitigate 
the adverse effects of the imbalance between previous and new data. To maintain fairness between old classes and 
new classes, BiC and WA \cite{wu2019large, zhao2020maintaining} correct the bias of the last fully connected layer by a linear model. To obtain the 
optimal exemplars for both the new class and the old class, EEIL \cite{liu2020mnemonics} proposes a novel and automatic framework mnemonics, 
where they parameterize exemplars and make them optimizable in an end-to-end manner. SDC \cite{yu2020semantic} proposes a new method called 
semantic drift compensation to deal with the drift of the data in incremental learning. PODNet and TPCIL \cite{douillard2020podnet, tao2020topology} replace the conventional 
distillation losses with an efficient spatial-based distillation-loss applied throughout the model and the 
topology-preserving constraint, respectively.

Recently, TOPIC \cite{tao2020few} presents a challenging but practical FSCIL problem. To address the 
problem, they propose a neural gas (NG) network to constrain feature space topologies for knowledge representation. 
We follow their FSCIL task settings. However, different from their work where a topology preservation framework is used 
to address insufficient samples in incremental training process, we adopt a non-training update mechanism to adapt 
few incremental samples based on extensible representation and explicit inter-class dependencies. 

\subsection{Few-shot Learning}
Few-shot learning is to learn a model that recognizes the class of new samples given few reference images. Generally, 
corresponding methods can be divided into three categories. Metric-based methods \cite{snell2017prototypical, sung2018learning, cai2014attribute, cai2012attribute} focus on the similarity metric function 
over the embeddings. Meta-based methods \cite{finn2017model, rusu2018meta, Zhu2020SelfSupervisedTF, he2020progressive} aims to learn a learning 
strategy to adjust well to new samples. Augmentation-based 
methods \cite{Li2020AdversarialFH} synthesize more data from the novel classes in different ways to facilitate standard learning. 

Most of these methods focus on the fast learning of the novel classes, while neglecting the recognition accuracy on the 
initial classes. To address this issue, \cite{gidaris2018dynamic} proposes a dynamic few-shot recognition system with an attention based few-shot classification weight 
generator. \cite{gidaris2019generating} introduces Graph Neural Network to capture the co-dependencies between
adjacent classes and promote the weight generation of new classes. These works aim to quickly adapt to novel classes from few training data while not forgetting the 
base classes \cite{Ren2019IncrementalFL} on which it was trained. Instead, we consider an extensible representation 
and global dependencies among classes at different sessions to maintain continuous stability during 
the incremental process.

\section{Problem Description}
The few-shot class-incremental learning (FSCIL) problem is defined as follows. Here we denote X, Y and Z as the training 
set, the label set and the test set, respectively. Our task is to train the model from a continuous data stream in a 
class-incremental form, \textit{i.e.,} training sets $X^{1},  X^{2}, \cdots X^{n}$, where samples of a set $X^{i}$ are from 
the label set $Y^{i}$, and $n$ represents the incremental session. It should be mentioned that all the incremental 
classes are disjoint, that is, $Y^{i}\cap Y^{j}=\varnothing (i\neq j)$. Except that there are sufficient samples in 
the first session $X^{1}$, only few samples (e.g., 5 samples) are available for each class in the subsequent sessions, which is consistent 
with the embodiment of FSCIL. To measure the performance of models in FSCIL task, we calculate 
the classification accuracy on the test set $Z^{i}$ at each session $i$. Different from the training set, the classes of 
the test set $Z^{i}$ are from all the seen label sets $Y^{1}\bigcup Y^{1}\cdots \bigcup Y^{i}$.  

\begin{table*}[t]
   \begin{center}
   \begin{tabular}{ccccccccccccc}
   \toprule[1pt]
   \toprule[1pt]
   \multirow{2}{*}{RESS}& \multirow{2}{*}{SPPR} & \multirow{2}{*}{FT} & \multicolumn{9}{c}{Sessions}                                          & \multirow{2}{*}{Average} \\ \cline{4-12}
                        &                           &                           & 1     & 2     & 3     & 4     & 5     & 6     & 7     & 8     & 9     &                          \\ 
                        \midrule
                        &                           & $\surd$                   & 64.10 & 56.49 & 52.0  & 46.24 & 42.36 & 37.86 & 36.43 & 33.99 & 32.30 & 44.64                    \\
                        & $\surd$                   &                           & 64.10 & 61.02 & 56.63 & 52.88 & 49.49 & 46.65 & 44.06 & 39.47 & 37.51 & 50.20                    \\
                        & $\surd$                   & $\surd$                   & 64.10 & 59.85 & 55.27 & 50.99 & 47.60 & 44.14 & 41.64 & 39.06 & 36.36 & 48.20                    \\
$\surd$                 &                           & $\surd$                   & 64.10 & 63.51 & 58.09 & 53.37 & 50.28 & 46.07 & 43.41 & 41.16 & 39.05 & 51.00                    \\
$\surd$                 & $\surd$                   &                           & \bm{$64.10$} & \bm{$66.10$} & \bm{$61.43$} & 57.33 & \bm{$53.72$} & 50.51 & 48.24 & 45.58 & 42.99 & 54.44                    \\ 
\midrule
$\surd$                 & $\surd$                   & $\surd$                   & \bm{$64.10$} & 65.86 & 61.36 & \bm{$57.34$} & 53.69 & \bm{$50.75$} & \bm{$48.58$} & \bm{$45.66$} & \bm{$43.25$} & \bm{$54.51$}                    \\ 
\bottomrule[1pt]
\bottomrule[1pt]
   \end{tabular}
   \end{center}
   \caption{Ablation study on CIFAR-100. All results are the average of multiple tests, and bold fonts represent the best results.}
   \label{fig:ablation}
\end{table*}

\section{Method}
We detail the incremental prototype learning scheme and its important components in this section. First of all, we 
demonstrate the paradigms of standard learning and our proposed incremental prototype learning, respectively. Then two 
core components random episode selection strategy and the self-promoted prototype refinement mechanism are introduced. 
Finally, we analyze the optimization flow of the overall pipeline and explain why it works well. 

\subsection{Standard Learning Paradigm} 
For the training process of the base classes (\textit{i.e.,} the first session) in incremental task, standard classification pipeline 
is adopted. In this case, the input of the model is only the query image Q to be predicted. Then a base feature 
extractor $f_{e}$ such as VGG \cite{Simonyan2015VeryDC} or ResNet \cite{he2016deep} parameterized by $\theta_{e}$ is utilized to learn the corresponding representation:
\begin{equation}
   R_{q} = f_{e}(Q;\theta_{e}).
\end{equation}
Finally, a certain metric $f_{m}$ parameterized by $\theta_{m}$ is used to measure the relationship between the representation and the learnable prototypes 
$\theta _{p}$ for all classes:
\begin{equation}
   \label{equa:overall}
   S = \operatorname{softmax}(f_{m}(R_{q}, \theta _{p};\theta_{m})).
\end{equation}
$f_{m}$ can represents a variety of classifiers, including the non-parametric ones (e.g., nearest-mean-of-exemplars classifier \cite{rebuffi2017icarl} and 
cosine classifier \cite{hou2019learning}) and the parametric ones (e.g., fully connected classifier \cite{Simonyan2015VeryDC}). Taking the cosine classifier as an example, the above 
formula can be written as:
\begin{equation}
   S_{i} = \frac{\exp (\eta ({\theta _{p}^{i}}^{T}\cdot R_{q})))}{\sum _{j}\exp (\eta ({\theta _{p}^{j}}^{T}\cdot R_{q})))},
\end{equation}
where $i$ is the calculated class, $\eta$ is the scale factor, and $\cdot$ represents the operation of inner product. In 
this case, if $\eta$ is learnable then $\theta_{m}$ refers to $\eta$, otherwise $\theta_{m}$ is empty. Our task is to 
randomly sample query images from the dataset, train and optimize $\theta $, thus minimizing the loss function $L$ under the 
supervision of target labels $T$:
\begin{equation}
   \theta _{\ast } = arg \min\limits_{\theta} L(S_{i}, T).
\end{equation}
Here $\theta $ includes above $\theta_{e}$, $\theta _{p}$ and $\theta _{m}$. In classification tasks, $L$ usually represents 
cross-entropy loss function.

\subsection{Incremental Prototype Learning}
\label{update}
As the representation obtained from standard learning lacks extensibility, we propose an incremental prototype learning 
scheme. There are two important components in the scheme as follows.

\textbf{Random Episode Selection.} We introduce the random episode selection strategy into the learning 
process and generate a N-way K-shot \cite{vinyals2016matching} incremental episode in each iteration. It enhances the extensibility of the feature representation by forcing gradients to adapt to different simulated 
incremental processes generated randomly. Unlike the few-shot task where the goal is only to identify the N classes, the simulated 
incremental process aims at identifying all seen classes with few samples of the N classes. Specificly, in addition to the 
above query image Q, the input of the model contains a randomly selected N-way K-shot collection C from the base training 
set $X^{1}$. As shown in Fig. \ref{fig:model}, in each iteration, N classes are randomly selected from the label 
space $Y^{1}$, and 
then K samples are selected for the feature extractor. The obtained embeddings are averaged for 
each class:
\begin{equation}
   R_{s} = \operatorname{mean}(f_{e}(C;\theta_{e})).
\end{equation}
Finally, these N classes are assumed not to have been seen before this iteration, so their corresponding  
prototypes will be eliminated. Mathematically,
\begin{equation}
   \theta _{p}^{N} = \mathbb{C}_{\left | Y^{1} \right |}^{\left | Y^{1} \right |-N}(\theta _{p}),
\end{equation}
where $\left | Y^{1} \right |$ represents the number of classes in label set $Y^{1}$, and $\mathbb{C}$ represents the mathematical operation that 
determines the possible arrangements in a collection of items (\textit{i.e.,} $\mathbb{C} _{n}^{m} = \frac{n!}{m!(n-m)!} $).
At this point, all the inputs and outputs of the model have been determined. Our goal is still to classify the query image Q given 
corresponding embeddings and prototypes, 
which is:
\begin{equation}
   S = P(R_{q}\mid R_{s}, \theta _{p}^{N}).
\end{equation}

\textbf{Dynamic Relation Projection.} To maintain the dependencies \cite{zhai2020deep, li2018visual, zhang2012attribute} of old classes and enhance the discrimination of the new classes, we propose a self-promoted prototype 
refinement mechanism $f_{u}$. In general, we obtain the refined prototypes $\theta _{p}$ under the guidance of relation matrix \cite{li2019online} 
between the embeddings of new classes and the prototypes of old classes: 
\begin{equation}
   \theta {_{p}}' = f_{u}(R_{s}, \theta _{p}^{N}; \theta_{u}).
\end{equation}
Specifically, the embeddings and the old prototypes are first transformed into a shared latent space,
\begin{align}
   T_{s} & = f_{t_{1}}(R{s}; \theta_{t_{1}}),\\
   T_{p} &= f_{t_{2}}(\theta _{p}^{N}; \theta_{t_{2}}),
\end{align}
where $f_{t_{1}}$ and $f_{t_{2}}$ represents a set of standard convolution block including a 1 $\times $ 1 convolution, 
a batch normalization layer and a ReLU activation layer. Then we calculate the cosine similarities 
between the old classes and the new classes in this space as follows:
\begin{align}
  T_{Y^{1}} &= Concat([T_{s}, T_{p}]),\\
  &Corr = T_{p} \cdot T_{Y^{1}}^{T}.
\end{align}
At this point, we obtain the relation matrix $Corr$ between the old and new classes, and use it as the transition coefficient 
of prototype refinement,
\begin{equation}
   \theta {_{p}}' = Corr^{T} \cdot \theta _{p}^{N}.
 \end{equation}
 Since the refinement mechanism not only explicitly considers the relation between the new and old classes, but is also 
 guided by the random selection process, the prototypes dynamically move toward maintaining existing 
 knowledge and enhancing the discrimination of new classes.

\subsection{Optimization}
To further understand the role of the above two parts, we analyze their impacts on the optimization process. 
Compared to standard learning, 
in addition to the extra parameters of the transform module that need to be optimized, the optimization directions of 
the feature representations and prototypes have also been significantly changed. Specifically, we integrate the 
optimization process of the parameter in the feature representation $\theta_{e}$ as follows:
\begin{equation}
\label{equa:optim1}
   \begin{aligned}
   S_{i} &= f_{m}(R_{q}, \theta {_{p}}';\theta_{m})\\
         &=f_{m}(f_{e}(Q;\theta_{e}), f_{u}(f_{e}(C;\theta_{e}), \theta _{p}^{N}; \theta_{u});\theta_{m}).
   \end{aligned}
\end{equation}
Under the new learning scheme, we not only learn a representation that is conducive to the classification 
of existing classes (the former $\theta_{e}$ in the above formula), but also encourage the network to reach an area in the 
parameter space where update of any class will be beneficial for subsequent incremental task (the latter $\theta_{e}$). 
For convenience, we omit the softmax operation in the formula.

For prototypes, it is no longer just supervised by classification labels without any other constraints. Compared to 
Eq. \ref{equa:overall}, the learnable prototypes perform joint optimization with the representation of selected 
collection S under the condition of satisfying mutual relation projection. Such a projection relation, as shown in the later 
visualization part, is well achieved in the optimization process.
\begin{align}
   \begin{aligned}
   S_{i} &= f_{m}(R_{q}, \theta {_{p}}';\theta_{m})\\
         &= f_{m}(f_{e}(Q;\theta_{e}), f_{u}(R_{s}, \mathbb{C}_{\left | Y^{1} \right |}^{\left | Y^{1} \right |-N}(\theta _{p}); \theta_{u});\theta_{m}).
   \end{aligned}
\end{align}

\begin{figure*}[t]
   \begin{center}
      \includegraphics[width=0.9\linewidth]{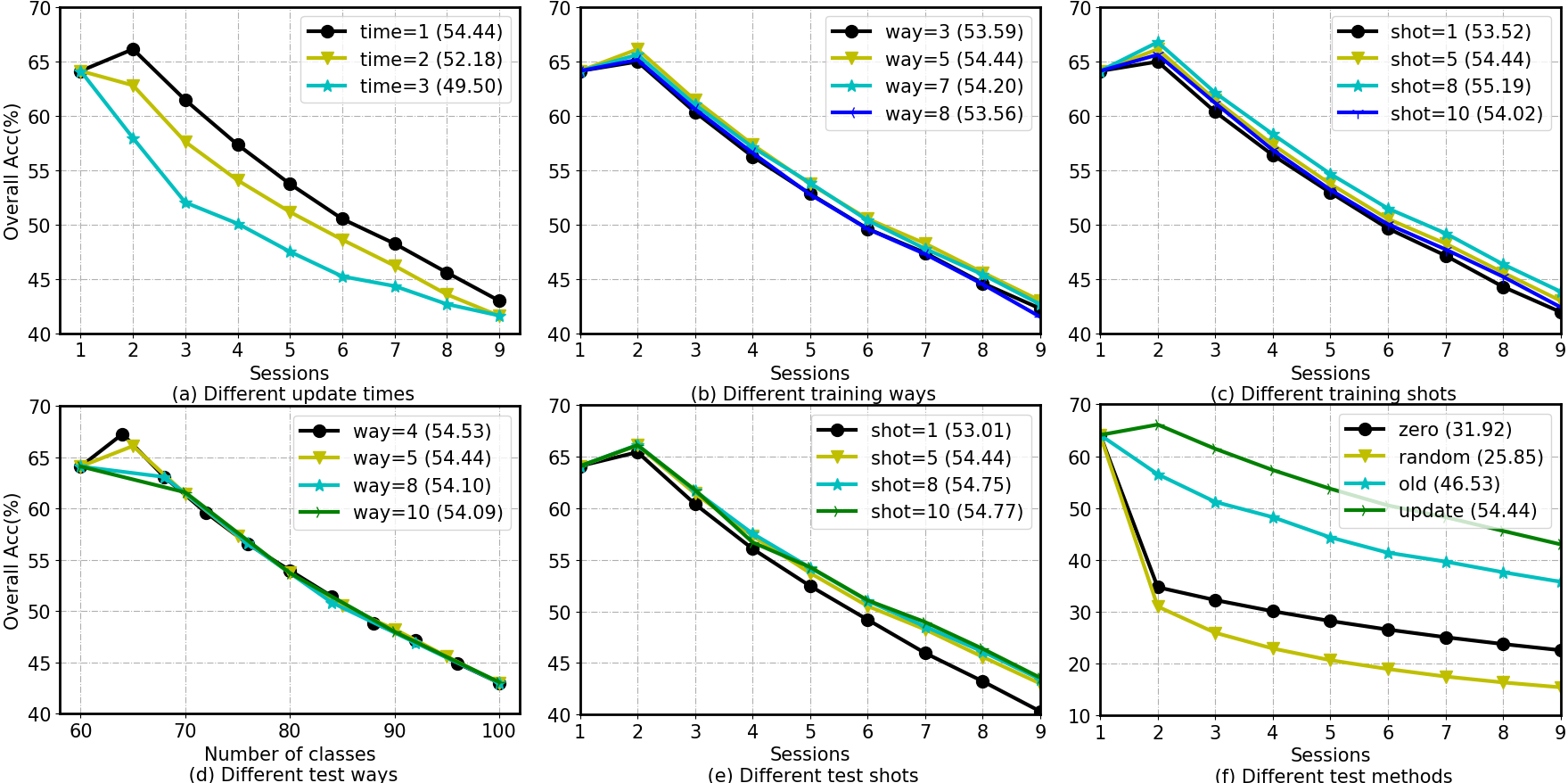}
   \end{center}
      \caption{Performance analysis under different conditions on CIFAR-100. (a) The initial extensible representations are obtained  
      through different times of updates. (b) During the random episode selection process, different numbers of 
      ways are chosen in N-way K-shot setting. (c) During the training process at session 1, different numbers of 
      shots are selected. (d) At session n (n$\geq $2), different numbers of incremental classes are used. (e) 
      At session n (n$\geq $2), different numbers of samples in each incremental class are used. (f) At  
      session n (n$\geq $2), different prototype update methods are used.}
   \label{fig:analysis}
\end{figure*}

\section{Experiment}
\subsection{Dataset and Settings}
\textbf{Dataset.} To evaluate the performance of our proposed method, we conduct comprehensive experiments on three datasets  
CIFAR-100 \cite{Krizhevsky2009LearningML}, MiniImageNet \cite{vinyals2016matching} and CUB200 \cite{Wah2011TheCB}. CIFAR-100 contains 60000 images of 32 $\times$ 32 size from 100 classes, 
and each class includes 500 training images and 100 test images. MiniImageNet contains 60000 images of 84 
$\times$ 84 size from ImageNet-1k \cite{Deng2009ImageNetAL}. Although it has the same number of classes and samples as CIFAR-100, 
its content is more complex and is valuable for the study of FSCIL. CUB200 is the most 
widely used benchmark for fine-grained image classification. The dataset covers 200 species of birds, 
including 5994 training images and 5794 test images. It provides more sessions and incremental classes to compare the sensitivity 
of different methods. For all the three datasets, we follow the same settings as \cite{tao2020few} including the division
of the datasets and incremental training samples. More details can be found in \cite{tao2020few}.

\textbf{Settings.} As adopted in \cite{tao2020few}, we use ResNet-18 as the backbone CNN.  
We show the classification accuracy at each session and the 
average accuracy as stated in most CIL work. 
\begin{figure}[t]
   \begin{center}
      \includegraphics[width=0.9\linewidth]{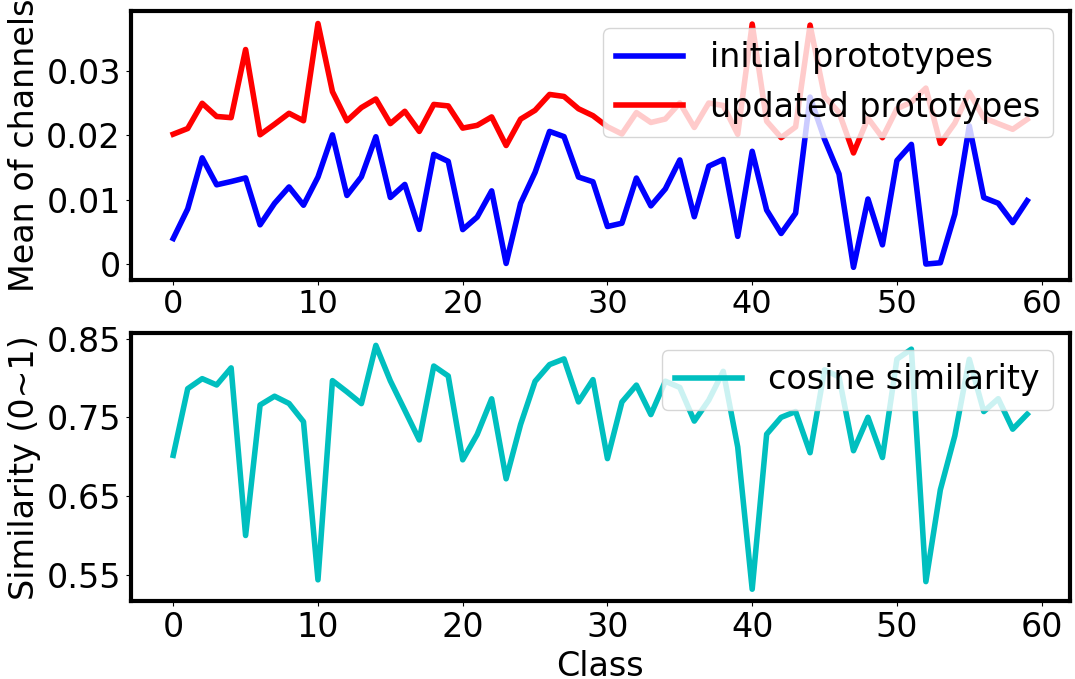}
   \end{center}
      \caption{Illustration of the role of the prototype refinement mechanism on CIFAR-100. The distribution of the prototypes before and after 
      update is shown in the upper row. The cosine similarity between the two is shown in the lower row.}
   \label{fig:similarity}
\end{figure}
To make a fair comparison, we achieve the same accuracy in base classes (\textit{i.e.,} session=1) for all the datasets denoted as $``Ours"$.
We also denote the best result without the constraint of the base classes as $``Ours^{\ast }"$.  
To reduce the error caused by random sampling of incremental samples, we select different samples to test 5 times 
and then take the average for each model. Since the average results are close to those with the same incremental samples as \cite{tao2020few}, 
we only provide the former in the text. The latter and the degree of dispersion in multiple tests are provided in the 
supplementary material. Our model uses the SGD optimizer during the 
training process. The initial learning rate is set to 0.02 and the attenuation rate is set to 0.0005. The model 
stops training after 70 epochs, and batch size is set to 128.

\begin{figure*}[t]
   \begin{center}
      \includegraphics[width=0.9\linewidth]{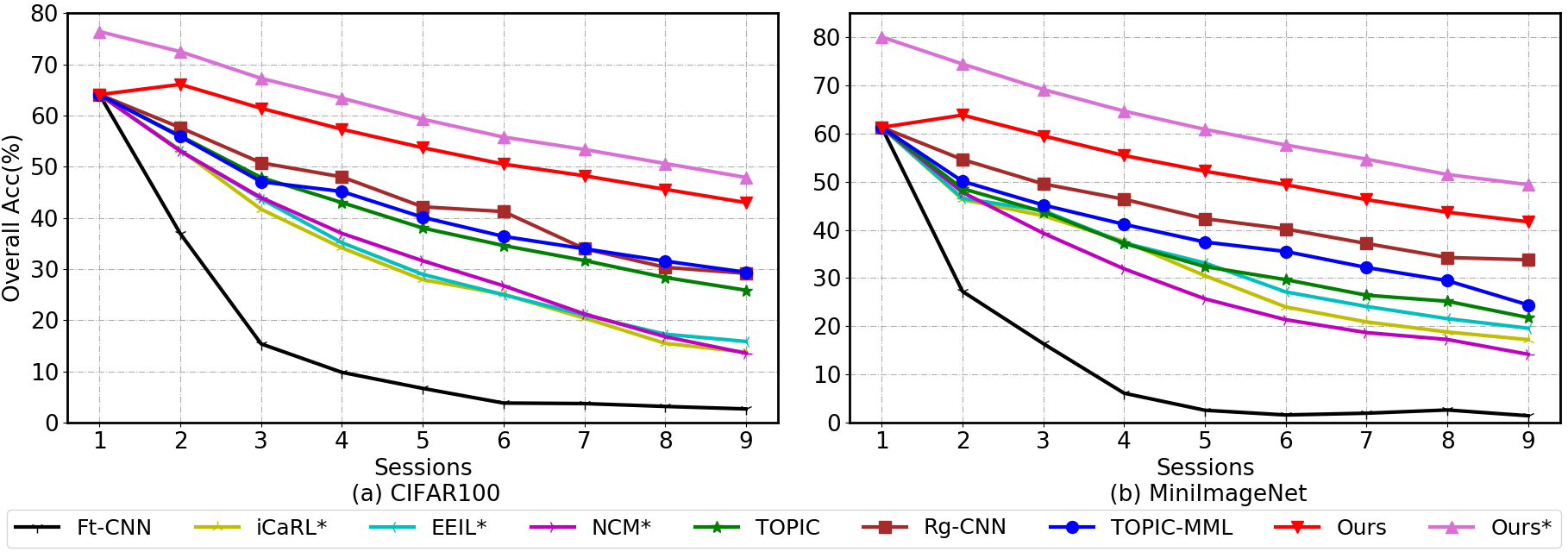}
   \end{center}
      \caption{Comparison of our classification results with other methods on CIFAR-100 and MiniImageNet. The abscissa represents 
      the incremental sessions and the ordinate represents the classification accuracy on the test set containing 
      all the seen classes.}
   \label{fig:compare}
\end{figure*}

\subsection{Ablation Study}
To prove the effectiveness of our proposed method, we conduct several ablation experiments on CIFAR-100. 
The performance of our scheme is mainly attributed to two prominent components: the random episode selection strategy (RESS)   
and the self-promoted prototype refinement mechanism (SPPR). To clarify the function of these two parts, we replace the 
extensible representation and the SPPR with the representation obtained by standard learning and the fine-tuning update method (FT), respectively. 

As can be seen from Table. \ref{fig:ablation} (the last three rows), the extensible representation brings a huge improvement 
in overall performance, about 5 percentage points on average. It demonstrates that extensible representation is far 
more useful than standard representation in FSCIL task. Without extensible representation, SPPR alone 
cannot play its role, and its performance drops by $4.24\% $. Due to the fact that SPPR enhances 
the expression ability of the feature representation, it brings an over 3 percent increment compared to 
fine-tuning. At the same time, we add the fine-tuning process to the overall scheme for observation. It can be seen 
that this step does not boost performance, which demonstrates the effectiveness of SPPR in updating the prototypes. 

\subsection{Analysis}
\textbf{The impact of the number of updates.} To explore the impact of the number of updates on extensible representation during training, we design the following 
experiments on CIFAR-100. We obtain different extensible representation by repeating the random selection process $n$ times and 
compare the test results. Take $n$ equal to 2 as an example. That is to say, the final 60 classes of prototypes will be obtained 
through two incremental updates from 50 classes of false base prototypes (5 classes each time). We try to further strengthen the extensible of the representation 
by increasing the iteration numbers of prototype updates. However, the results show that with the number of iterations increases, 
the corresponding accuracy drops as in Fig. \ref{fig:analysis} (a). We argue that the difficulty of training may also increase sharply through 
multiple iterations, so the obtained representation does not benefit from it. 

\begin{figure*}[t]
   \begin{center}
      \includegraphics[width=0.9\linewidth]{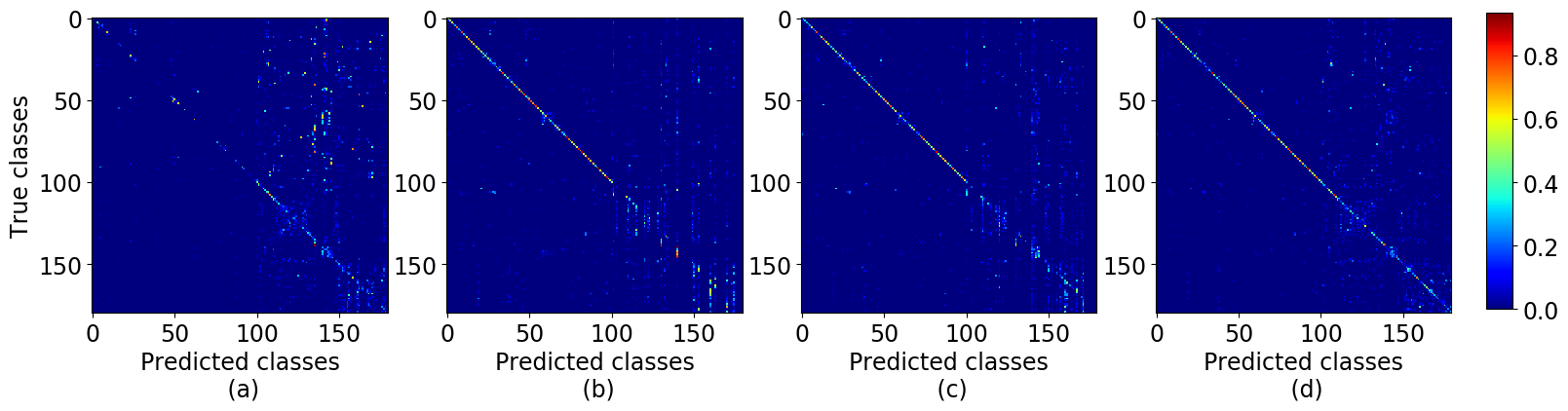}
   \end{center}
      \caption{Confusion matrix of four different variations on CUB200. (a) Ft-CNN. (b) Fixed representation with distill loss function. 
      (c) Extensible representation with distill loss function. 
      (d) Our method.}
   \label{fig:confusion}
\end{figure*}

\begin{table*}[t]
   \begin{center}
   \begin{tabular}{cccccccccccc}
      \toprule[1pt]
      \toprule[1pt]
   \multirow{2}{*}{Method} & \multicolumn{11}{c}{Sessions}                                                                                                                                                            \\ \cline{2-12} 
                           & 1              & 2              & 3              & 4              & 5              & 6              & 7              & 8              & 9              & 10             & 11             \\ 
   Ft-CNN                  & 68.68          & 44.81          & 32.26          & 25.83          & 25.62          & 25.22          & 20.84          & 16.77          & 18.82          & 18.25          & 17.18          \\ 
   \midrule
   iCaRL$^{\ast }$                   & 68.68          & 52.65          & 48.61          & 44.16          & 36.62          & 29.52          & 27.83          & 26.26          & 24.01          & 23.89          & 21.16          \\
   EEIL$^{\ast }$                   & 68.68          & 53.63          & 47.91          & 44.20          & 36.30          & 27.46          & 25.93          & 24.70          & 23.95          & 24.13          & 22.11          \\
   NCM$^{\ast }$                     & 68.68          & 57.12          & 44.21          & 28.78          & 26.71          & 25.66          & 24.62          & 21.52          & 20.12          & 20.06          & 19.87          \\ 
   \midrule
   TOPIC                   & 68.68          & 61.01          & 55.35          & 50.01          & 42.42          & 39.07          & 35.47          & 32.87          & 30.04          & 25.91          & 24.85          \\
   TOPIC-MML               & 68.68          & \bm{$62.49$}          & 54.81          & 49.99          & 45.25          & 41.40          & 38.35          & 35.36          & 32.22          & 28.31          & 26.28          \\ 
   \midrule
   Ours                    & \bm{$68.68$} & $61.85$ & \bm{$57.43$} & \bm{$52.68$} & \bm{$50.19$} & \bm{$46.88$} & \bm{$44.65$} & \bm{$43.07$} & \bm{$40.17$} &\bm{$39.63$} &\bm{$37.33$} \\ 
   \bottomrule[1pt]
	\bottomrule[1pt]
   \end{tabular}
   \end{center}
   \caption{Comparison of our classification resullts with other methods on CUB200.}
   \label{table:compare}
\end{table*}

\textbf{The sensitivity of the number of ways and shots during training.} To verify the impact of selection classes and sample numbers during training, we train multiple models for comparison. As 
can be seen in Fig. \ref{fig:analysis} (b) and (c), the training processes of different ways and different shots achieve 
similar result. The performance of 1-shot learning is slightly worse than other conditions, and the performance 
also drops slightly when the number of ways exceeds 5. It suggests that our learning strategy is not sensitive to the change 
of the number of episodic classes and samples. For convenience, we set the fixed incremental classes (N=5, K=5) for all 
the training process in this paper.

\textbf{The sensitivity of the number of ways and shots during test.} To explore whether the obtained representation is adapted to different test settings, we show the test results with different 
numbers of incremental classes and samples. As can be seen in Fig. \ref{fig:analysis} (d), the feature representation 
obtained by 5-way training can achieve almost the same curve in the case of different test sessions, 
which further demonstrates the robustness of our proposed method. In Fig. \ref{fig:analysis} (e), when the shot of the test images 
is reduced to 1, the final result has a significant drop. However, when the shot of images 
increase to 5 and 10, it makes nearly no difference. We think it is because we obtain the class embeddings directly by 
averaging all the shots, which will not benefit from the increase of shots. Finally, we tested different update methods, 
namely setting new prototypes to zero, random numbers and keeping old prototypes unchanged. It can be 
seen in Fig. \ref{fig:analysis} (f) that our method achieves the best results.

\subsection{Visualization}
To verify the role of the self-promoted prototype refinement mechanism in this task, we show the following visualization results. 
As shown in Fig. \ref{fig:similarity}, the initial prototypes of 60 base classes from CIFAR-100 dataset are averaged in the 
feature dimension and visualized 
in blue. Then 25 samples from 5 random classes are chosen and utilized to update the prototypes as Section. \ref{update} 
states. It can be seen that the updated prototypes in red are close to the initial ones in both value and trend. And their high 
cosine similarity demonstrates that almost every prototype has the same distribution in the feature dimensions before and 
after the update. 

\subsection{Comparison with SOTA}
To better assess the overall performance of our scheme, we compare it to the state-of-the-art methods of FSCIL  
(TOPIC and TOPIC-MML \cite{tao2020few}) and some classical methods of CIL (iCARL$^{\ast }$ \cite{rebuffi2017icarl}, 
EEIL$^{\ast }$ \cite{castro2018end} and NCM$^{\ast }$ \cite{hou2019learning}). The following asterisk represents 
the result of applying corresponding CIL 
methods to the FSCIL task. In addition, we set the fine-tuning method (Ft-CNN) as the baseline, and adopt some 
regularization techniques (\textit{i.e.,} weight regularization, data augmentation and distillation) on this basis (Rg-CNN) for 
comparison.

\textbf{CIFAR-100 and MiniImageNet.} It can be seen in Fig. \ref{fig:compare} that under the average accuracy metric, 
our method surpasses the SOTA method over 13 percentage on CIFAR-100 
dataset, and yields 17 percentage improvement on MiniImageNet dataset. At the same time, our incremental classification results 
is higher than other methods at all sessions, and the attenuation is also slower. To make a fair comparison, we provide the results 
under the same accuracy of base classes, which are far below the best accuracy of ResNet-18 on these two datasets. This is why the results 
of the second session are even higher than the first session. The increment in the accuracy of the old classes benefits from the reverse 
effect of the relation projection. To manifest the real function of our method, we show the accuracy curve of the best 
result without the constraint of the base classes in violet line. It can be seen that in this case we get the best 
extensible representation and classification result. 

To provide further insight into the behaviors of different methods, we compare their confusion matrix. As shown in 
Fig. \ref{fig:confusion}, fine-tuning tends to classify all the samples into the incremental classes due to 
overfitting. Typical incremental methods (\textit{i.e.,} fixed or extensible representation followed by different distill loss 
functions) often make mistakes when distinguishing newly incremental classes because of the 
lack of discriminative prototypes. The confusion matrix of our method suggests the superiority of both the representation 
and prototypes in all classes.

\textbf{CUB200.} As shown in Table \ref{table:compare}, we achieve over 11 percentage improvement compared to 
the SOTA method. Since the 
number of incremental classes is twice than each of the above datasets, the forgetting rate at each session is much higher. 
It can be seen that the difficulty increases as the number of incremental classes and sessions increases. Different from 
that on above datasets, the initial classification accuracy (68.68$\% $) is close to the best result ($``Ours^{\ast }"$), so we only report 
one result ($``Ours"$) on this dataset.

\section{Conclusion}
In this paper, a novel incremental prototype learning scheme is proposed for FSCIL task. A 
random episode selection strategy is firstly proposed to enhance the extensibility and optimization capability of feature representation, and then all the   
prototypes are reorganized with a self-promoted prototype refinement mechanism. Consequently, our method incorporates incremental classes with few 
samples into recognition. Experimental results show that our model is superior in both performance and adaptability with 
respect to SOTA methods.

\noindent\textbf{Acknowledgments.} This work was supported by the National Key R\&D Program of China under Grand 2020AAA0105702, National Natural 
Science Foundation of China (NSFC) under Grants 61872327 and U19B2038, the Fundamental Research Funds for the Central 
Universities under Grant WK2380000001 as well as Huawei Technologies Co., Ltd.
{\small
\bibliographystyle{ieee_fullname}
\bibliography{egbib}
}

\end{document}